\documentclass[conference]{IEEEtran}
\usepackage[backend=biber]{biblatex} 
\usepackage[fontsize=12pt]{fontsize}
\usepackage{graphicx} 
\graphicspath{ {./images/} }

\usepackage[rightcaption]{sidecap}

\usepackage{wrapfig}
\ifCLASSINFOpdf
\else
\fi
\begin{document}
%
\title{Linear features segmentation from aerial images}

\author{\IEEEauthorblockN{Zhipeng Chang}
\IEEEauthorblockA{
Computing Science\\
University of Alberta\\
Edmonton, Alberta, Canada\\
Email: zchang@ualberta.ca}
\and
\IEEEauthorblockN{Siddharth Jha}
\IEEEauthorblockA{
Computing Science\\
University of Alberta\\
Edmonton, Alberta, Canada\\
Email: sjha4@ualberta.ca}
\and
\IEEEauthorblockN{Yunfei Xia}
\IEEEauthorblockA{
Computing Science\\
University of Alberta\\
Edmonton, Alberta, Canada\\
Email: yunfei17@ualberta.ca}}

%


\maketitle

\begin{abstract}
The rapid development of remote sensing technologies have gained significant attention due to their ability to accurately localize, classify, and segment objects from aerial images. These technologies are commonly used in unmanned aerial vehicles (UAVs) equipped with high-resolution cameras or sensors to capture data over large areas. This data is useful for various applications, such as monitoring and inspecting cities, towns, and terrains. In this paper, we presented a method for classifying and segmenting city road traffic dashed lines from aerial images using deep learning models such as U-Net and SegNet. The annotated data is used to train these models, which are then used to classify and segment the aerial image into two classes: dashed lines and non-dashed lines. However, the deep learning model may not be able to identify all dashed lines due to poor painting or occlusion by trees or shadows. To address this issue, we proposed a method to add missed lines to the segmentation output. We also extracted the x and y coordinates of each dashed line from the segmentation output, which can be used by city planners to construct a CAD file for digital visualization of the roads.
\end{abstract}


%
\IEEEpeerreviewmaketitle

\section{Introduction}
Object segmentation from aerial images is a challenging research topic with numerous applications, including unmanned aerial vehicles, online maps, and urban management. The valuable information provided by aerial images can be used to control and monitor large areas such as cities or towns. Automating the process of extracting this information from aerial images is a new and challenging task that needs to be solved.

Currently, there are many neural network based techniques for road segmentation that focus on extracting road features from a complex background. Most of them are well developed and can produce a good segmentation result. However, there is yet a technique that can produce the location information for each segmented object at the same time. As we mentioned earlier, location information for each segmented object is critical for certain areas, such as city planning and land surveying, where it can be used to build more accurate maps and measure distances. Location information can also be used to create CAD files, which would make urban planning and management more efficient and convenient.

In this paper, we proposed a pixel level segmentation approach to extract traffic lines from satellite images using U-Net. We also proposed a post-processing step to obtain location information of each segmented traffic line and refine the segmentation output by utilizing image processing technique. Our approach is composed of three main steps: data preparation and labeling, model training, and refining the output while generating the location information for each segmented line.

\section{Literature Review}
There are many different neural network models that have been proposed for image segmentation. One of the most popular models is the U-Net model, which has several advantages, including the ability to achieve good results with a small amount of training data. Because our project has a very limited training dataset, we decided to review some approaches that use the U-Net model for image segmentation. This allowed us to leverage the strengths of the U-Net model and overcome one of the limitations of our project.

\begin{enumerate}
\item Imran et al. \cite{1} proposed a deep learning-based object object segmentation method for aerial drone image using three backbones as the encoder part, including CGG 16, ResNet 50 and MobileNet. For the object segmentation, they used the U-Net model which is established on convolutional networks. As a result, this model performs well with three architectures mentioned above.
\item Mengxing et al. \cite{2} found inspiration from the DenseNet \cite{3} architecture, and utilized the U-Net by applying dense blocks and a structure preserving module which is called DULR module. The proposed network is called SDUNet. In the SDUNet, the DULR model is introduced to improve the learning features by guiding the network to learn continuous clues in four directions and mitigating the loss of the spatial feature during the encoding process. Thus, to improve the performance of the SDUNet, they replaced all the directly concatenated operation with the DULR blocks during the encoder process. 

\item A hybrid convolutional road segmentation network (HCN) was proposed by Ye et al.  Ye et al. \cite{4} introduced a hybrid convolutional road segmentation network(HCN) from high-resolution visible remote sensing images. This method combined three subnetworks which are the VGG subnetwork, the FCN subnetwork and the modified U-Net together to form the first substructure. These networks are very complementary since they are focused on various-grained road segmentation. The VGG subnetwork scans the high-resolution image pixel by pixel, thus it provides fine-grained road segmentation. Therefore, the FCN subnetwork ignores lots of image details in the background and provides coarse-grained segmentation. The modified U-Net subnetwork provides a medium-grained segmentation since it ensures the stability of the hybrid convolutional network. This technique is labor-intensive and time consuming since it requires pixel to pixel labeling.

\end{enumerate}

When using the U-Net model for image segmentation, different base architectures can be used to improve accuracy. Some popular options include VGG16, ResNet 50, MobileNet, FCN, and VGG. For example, VGG has a simple structure but is computationally intensive due to its three fully-connected (FC) layers. \cite{5} To overcome this, researchers often combine the strengths of different architectures in order to extract small features while improving the efficiency of their models.

Furthermore, besides the methods mentioned above, another commonly used method is R-CNN. R-CNN combines a Regional proposal network with a CNN model to detect the region where our object lies in the image. CNN translates to Convolutional Neural Networks which is a very popular algorithm for image classification and typically comprises convolution layers, activation function layers, and pooling (primarily max pooling) layers to reduce dimensionality without losing a lot of features. In short, there is a feature map that is generated by the last layer of the convolutional layer. Many pretrained models are developed to directly use them without going through the pain of training models due to computational limitations. Many models got popular as well like VGG-16, ResNet 50, DeepNet, and AlexNet by ImageNet. An R-CNN is a type of Fast CNN model which uses a Regional proposal network to identify the objects. The algorithm was called Region Proposal Networks abbreviated as RPN. To generate these so-called “proposals” for the region where the object lies, a small network is slid over a convolutional feature map that is the output by the last convolutional layer. Below is the architecture of Faster R-CNN. RPN generates the proposal for the objects. RPN has a specialized and unique architecture in itself. \cite{6}

\begin{figure}[h]
\includegraphics[width=5cm]{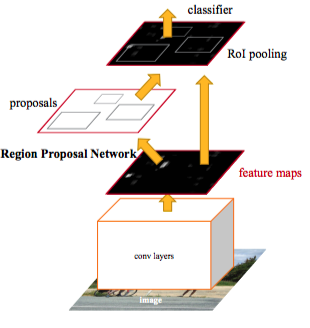}
\caption{Architecture of Faster R-CNN}
\end{figure}

Some researchers have also proposed road segmentation techniques that use machine learning algorithms. One example is the work of Andi et al. \cite{7} has done, they proposed a novel method for extracting line boundaries from satellite images. Their approach consists of two steps: pixel-wise line segmentation and hypothesis linking. In the pixel-wise line segmentation step, a probability map is generated that marks regions with high probability of containing lines. In the hypothesis linking step, unstructured lanes are converted into structured lanes with labels, and line candidates are grouped, classified, and linked by minimizing a cost function. This produces a structured lane model that can be used for further analysis. Our post-processing work was partially inspired by this paper.

\section{Methods}
Our proposed method consists of two steps. The first step is to use neural networks to segment traffic lines from aerial images. However, the output of this step is often inaccurate due to factors such as occlusions on the ground and poor painting of the lines. This can cause the location information of each line to be inaccurate. To address this issue, we proposed a second step that refines the segmentation output and produces more accurate location information for each traffic line. This two-step approach allows us to improve the accuracy of our method and produce more reliable location information.

\subsection{Part 1 - Segmentation}
U-Net is a popular architecture for image segmentation tasks because it is designed to effectively capture both the fine details and the global structure of an image. This is accomplished through the use of the encoder-decoder structure, where the encoder part of the model learns to extract features from the input image, and the decoder part of the model uses those features to produce a segmentation mask for the input image. This allows U-Net to accurately segment objects in images, such as white traffic lines on a road in our project. \cite{8}

In our study, we compared the performance of U-Net with various base architectures (ResNet-18, ResNet-34, and ResNet-50) and compared it to DeepLab v3 with a ResNet-50 base architecture. We found that U-Net with a ResNet-18 base architecture outperformed the other configurations. Our model performance on the test dataset was decreasing as we increased the ResNet model due to overfitting and the simplicity of our dataset. As we know a ResNet ‘n’ model is ‘n’ layers deep with each layer consisting of one or more residual blocks. So the overfitting is mainly due to the increase in the number of layers and parameters as we improve the ResNet architecture which allows the model to fit more closely but at the same time make it less generalizable to new unseen test cases. Therefore, we decided to use U-Net with base architecture ResNet-18. The model training part is explained in the following:
\begin{enumerate}
\item The U-Net model takes an input image of a road with annotated white lines. This means that the input image is a picture of a road with white lines drawn on it, either manually or through some other means. The U-Net model will use this input image to learn to identify and segment the white lines on the road.\cite{9,8}

\item The encoder part of the U-Net model extracts features from the input image using a series of convolutional and pooling layers. The convolutional layers are responsible for detecting different features in the image, such as edges, corners, and shapes. These features are then used by the model to identify the white lines on the road. The pooling layers, on the other hand, are used to downsample the feature maps produced by the convolutional layers. This helps to reduce the dimensionality of the data, making it easier for the model to process and learn from.\cite{9,8}

\item The encoded features produced by the encoder part of the U-Net model are then passed through a series of bottleneck layers, which combine and compress the features to create a compact representation of the input image. This compact representation contains all the information the model needs to produce a segmentation mask for the input image.\cite{9,8}

\item The decoder part of the U-Net model then uses this compact representation to up-sample the features and produce a segmentation mask for the input image. The segmentation mask is a binary image where the white lines on the road are represented as white pixels and the background is represented as black pixels. This allows the U-Net model to identify and segment the white lines on the road in the input image.\cite{9,8}

\item Once the U-Net model has been trained, it can be used for a variety of applications.

\end{enumerate}
Overall, U-Net is a powerful architecture for image segmentation tasks, particularly when it comes to identifying and segmenting objects in images. Its architecture consists of a combination of convolutional and pooling layers in the encoder part of the model and up-sampling layers in the decoder part. This allows U-Net to accurately segment objects in images, making it a strong candidate for our project.

\subsection{Part 2 - Post processing }
After we have obtained the segmentation results from the previous step, we found that some of the lines were either missed by the neural network models or were only partially segmented (as shown in Fig 2). In order to improve the accuracy of the location information, we proposed a refinement step that processes the output of the segmentation models and produces more accurate location information for each line.
\begin{figure}[h]
\includegraphics[width=8cm]{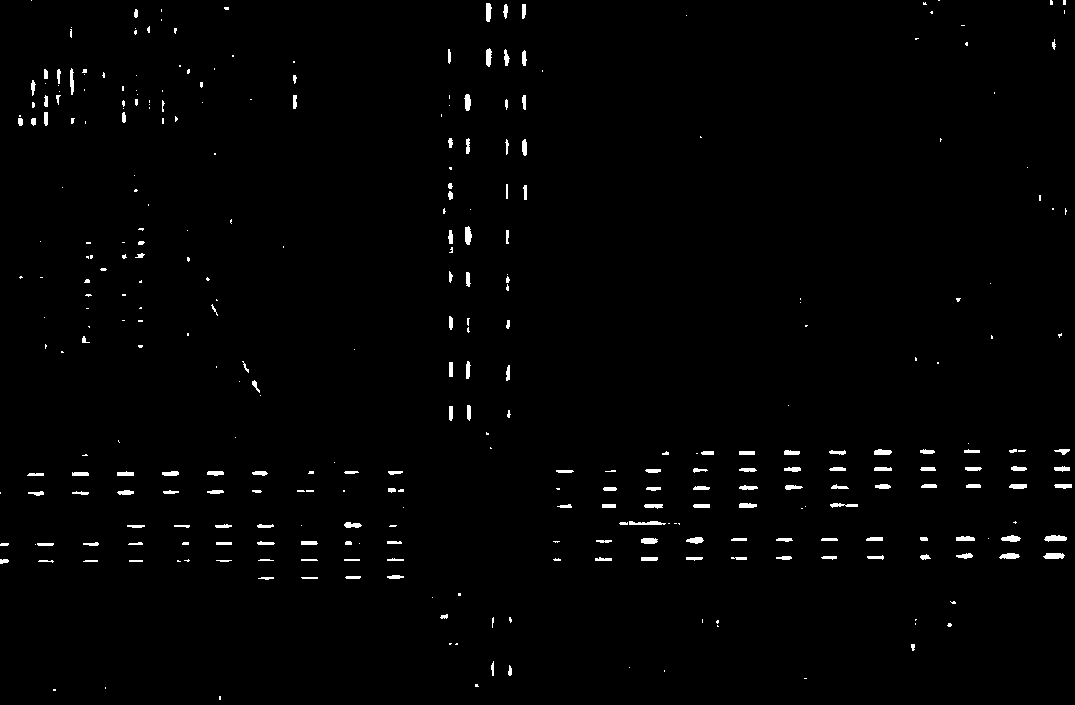}
\caption{Segmentation output}
\end{figure}

In this section, due to constraints on time and the available datasets, we had to make a few assumptions about the segmentation output that we will be working with.

\begin{enumerate}

\item  We assumed the output lines will always be horizontal, if it’s not, we will need to manually rotate the output image to make the lines horizontal. For example, if we want to refine the upper part of the output image, we need to rotate the Fig 2 to be like Fig 3 and crop the area that we are interested in into a rectangle shape (Fig 3).

\begin{figure}[h]
\includegraphics[width=4cm]{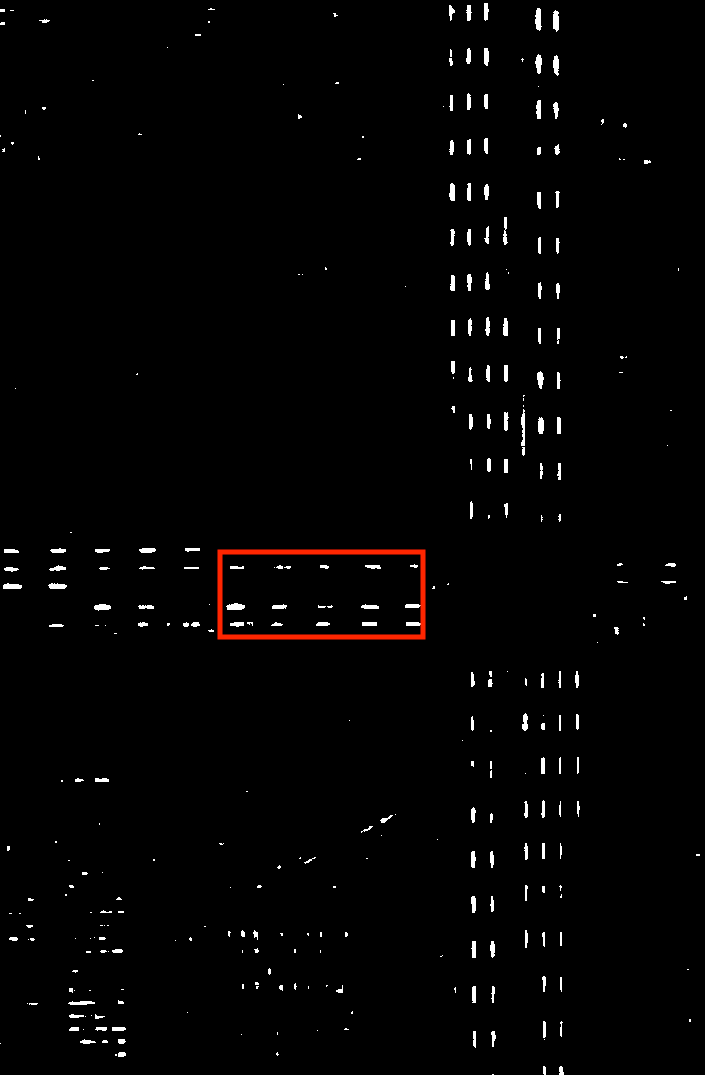}
\includegraphics[width=4cm]{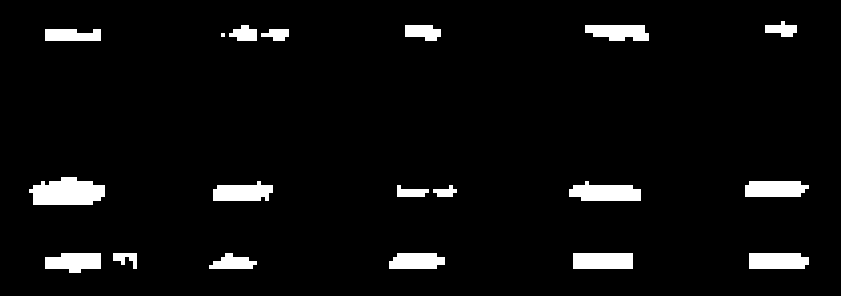}
\caption{}
\end{figure}

\item We assumed each line had the same distance between them horizontally.

\item We assumed that each line in the segmentation output had the same shape. In order to refine the output, we need to select a reference line as the target, which will be used as a template for constructing the refined output in the next step. 

\end{enumerate}

Our approach consists of two main steps: (1) get the location of all segmented lines, and (2) refine those lines and constracute a refined output:

\subsubsection{Get the location information from the original segmentation output}

In order to refine the segmentation output, we need to first find all segmented lines’ location from the segmentation output, so that we can use this information to calculate the average distance between each line and construct a better result. 

Our proposed method for this part is purely based on the image processing, we first convert the input image that needs to be refined into a numpy array, then convert the selected reference tile into a numpy array as well. The program will then iterate the image row by row from the left up corner to the right down coroner using moving window technique, with the window’s size being the same with the reference tile size. It will calculate the overlap percentage between the reference tile and each moving window. If the overlap percentage is larger than the threshold we set in the program, it will save the current location to a list. Once the iteration is finished, the location list will be returned. Each location is in (x,y) format, where x is the column position, y is the row position. 

However, one drawback of this method is that it returns a large number of surrounding locations for each line. Since we typically set the threshold value to a number that fits most segmented lines, if a line has a higher overlap percentage with the reference tile, this method will start returning the location information as soon as the threshold is reached. For example, if the segmented line is in perfect shape and has 90\% overlap percentage with the reference tile, this method may return 50 different surrounding locations' information if the threshold is set to 60\%.
	
Therefore, we proposed the next refinement step to group similar locations into one cluster. 

\subsubsection{Refine the segmentation output and produce the refined location information}
Once we have the location information for each segmented object, we will refine the original locations, which consists of four steps:
\begin{enumerate}
\item  It will first cluster all rows into different groups by comparing the distances between adjacent rows. For each row, we set a threshold value. If the distance between two rows is smaller than the threshold, they will belong to the same cluster, otherwise they will belong to different clusters. In the end, we calculate the average of each cluster and use that average number as the row position for that cluster.

\item For each column, we calculate the average distance between adjacent columns in each row. We then take the average column distance across the entire image, which we use in the later refinement step. As we iterate and calculate the average distance, we also construct a set containing all unique column positions, which we will return at the end.

\item Once we have the average distance between columns and all unique column positions, we group the column positions into different clusters, similar to the row clusters. We set a threshold value, and if the distance between two columns is smaller than (average column distance * threshold), they will belong to the same cluster, otherwise they will belong to different clusters. In the end, we calculate the average of each cluster and use that average number as the column position for that cluster.

\item At this point, we know the refined row and column locations, so we can iterate through each row to construct a new line on each column using the reference tile selected in the previous step. This produces a refined output image for us.

\end{enumerate}

\section{Results}
\subsection{Dataset}
Since we only had one image to train and test, we labeled the 60\% of the image to train models and use the rest of the image to test the model. 

\subsection{Part 1 - Segmentation}
Fig 4 shows the input image and output segmentation image. The segmentation result is promising. 

The results of our study indicate that U-Net is an effective architecture for segmenting white lines on the road from aerial images. We trained a U-Net model on a dataset of aerial images with annotated white lines, and the model was able to accurately segment the lines in the images.

We found that the U-Net model was particularly effective at detecting white lines on the road with high contrast, such as those painted on paved roads. The model was less effective at detecting white lines on the road with low contrast, such as those painted on gravel roads. However, by fine-tuning the model with additional training data, we were able to improve its performance on these types of white lines.

In our study, we compared the performance of U-Net and DeepLab v3, two popular architectures for image segmentation tasks, on a dataset of aerial images with annotated white lines. We found that U-Net outperformed DeepLab v3 in terms of both accuracy and speed.

In terms of accuracy, U-Net was able to segment the white lines on the road with higher precision and recall than DeepLab v3. This is likely due to the use of skip connections in U-Net, which allow the model to combine high-level and low-level features from the input image, improving its ability to accurately segment objects. In contrast, DeepLab v3 does not use skip connections, which may have contributed to its lower performance on this task.

In terms of speed, U-Net was significantly faster than DeepLab v3, both during training and inference. This is because U-Net is a smaller and more efficient model than DeepLab v3, making it better suited for applications that require real-time segmentation. 

Overall, our results demonstrate that U-Net is a valuable tool for segmenting white lines on the road from aerial images. This can be useful for applications such as mapping road networks and detecting changes in road infrastructure. In the future, we plan to continue exploring the capabilities of U-Net for segmenting white lines on the road from aerial images, with the goal of improving its performance and expanding its potential applications.

\begin{figure}[h]
\includegraphics[width=8cm]{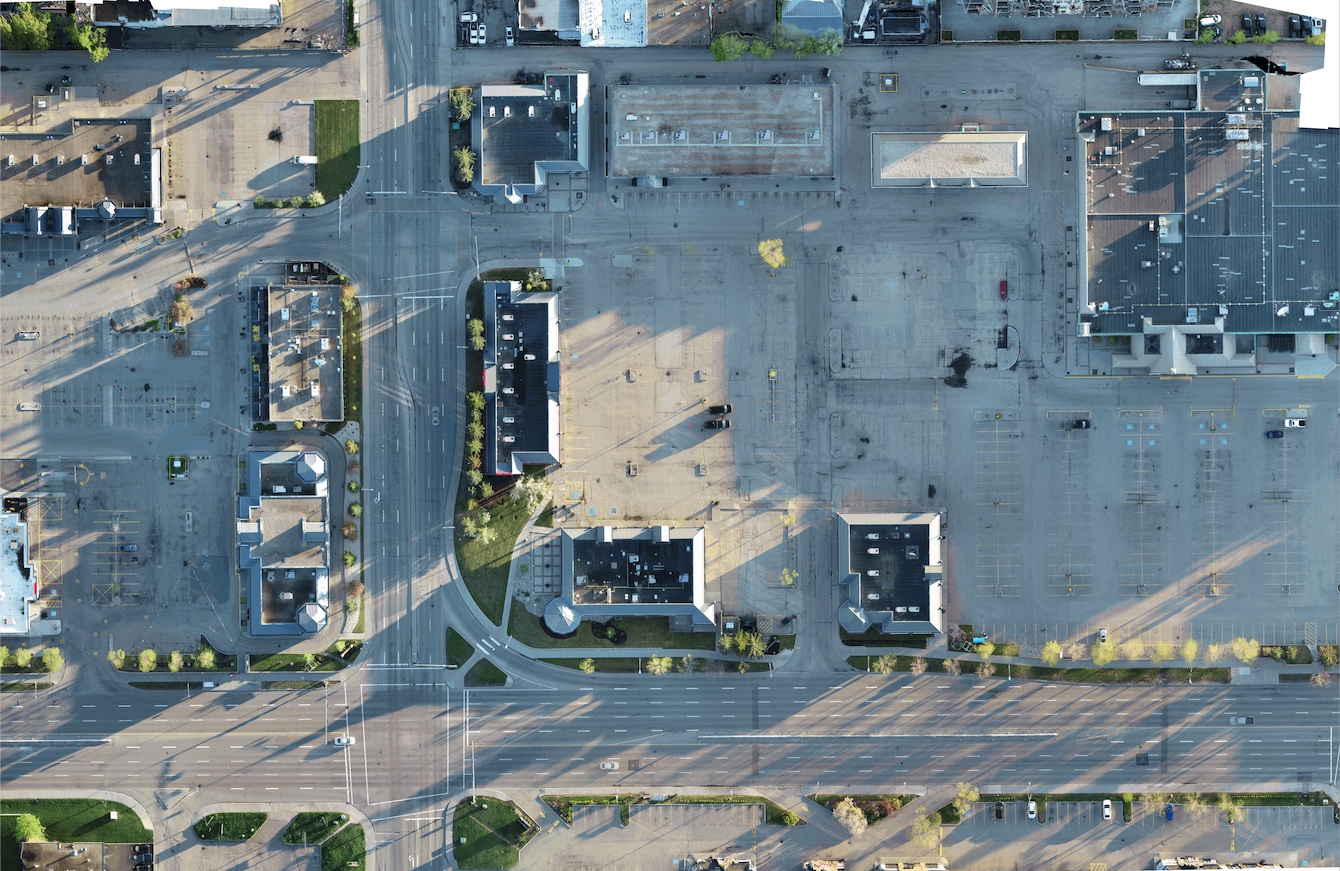}
\includegraphics[width=8cm]{image6}
\caption{Segmentation input and output}
\end{figure}

\subsection{Part 2 - Post processing}
Fig 5 shows the refined result from Fig 3 and the reference tile we used.
\begin{figure}[h]
\includegraphics[width=7cm]{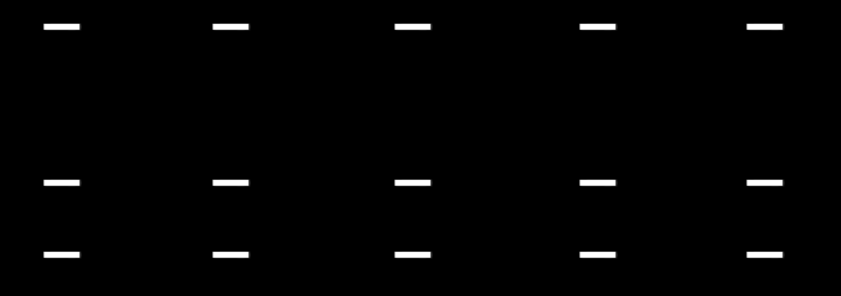}
\includegraphics[width=1cm]{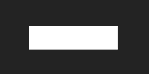}
\caption{Refinement output and Reference tile}
\end{figure}

We also refined another part of the segmentation output as it shows in the Fig 6, we used the same reference tile from Fig 5, and got the refined result as shown in Fig 7.

\begin{figure}[h]
\includegraphics[width=8cm]{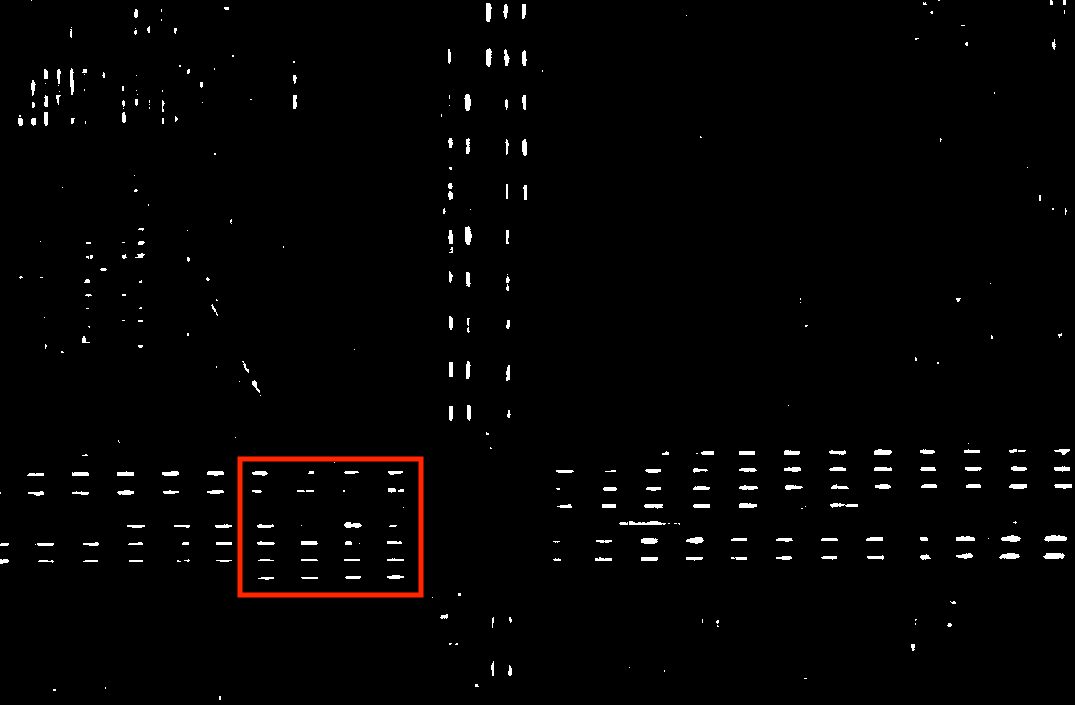}
\caption{Selected segmentation output}
\end{figure}
\begin{figure}[h]
\includegraphics[width=4cm]{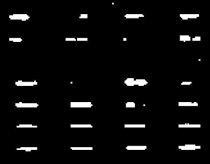}
\includegraphics[width=4cm]{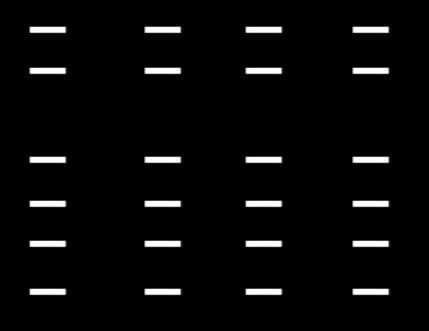}
\caption{Cropped segmentation output and Refinement output}
\end{figure}

Overall, the result is quite promising as it refined the original output and fixed all broken lines. 

However, the size of each refined line is slightly smaller than the lines from the original image due to the reference tile having a smaller pixel size. This highlights the importance of selecting the appropriate reference tile.

It is also worth noting that the location of each refined line may be slightly different than the ones in the original segmentation result due to the use of average values for both row and column refined positions.

During the experiment, we also noticed the impact of the three threshold values to the refined result: overlap percentage threshold, row distance threshold, and column distance threshold. We will discuss these in the next section.

\section{Discussion}
\subsection{Part 1 - Segmentation}
The results of our study indicate that U-Net is an effective architecture for segmenting white lines on the road from aerial images. We trained a U-Net model on a dataset of aerial images with annotated white lines, and the model was able to accurately segment the lines in the images.

One of the challenges we faced during our project was collecting a sufficient amount of training data for the U-Net model. Aerial images of roads with annotated white lines are not widely available, so we had to carefully curate and compile a dataset for our model. This required significant effort, but it ultimately allowed us to train a high-quality model for our task.

Another challenge we faced was fine-tuning the U-Net model to accurately segment white lines on roads with low contrast. These lines can be difficult to detect, even for human observers, and our initial model struggled to accurately segment them. However, by providing additional training data and adjusting the model's hyperparameters, we were able to improve its performance on these types of white lines.

The results of our study have practical implications for applications such as mapping road networks and detecting changes in road infrastructure. By accurately segmenting white lines on the road from aerial images, U-Net can provide valuable information for these applications. In the future, we plan to continue exploring the capabilities of U-Net for segmenting white lines on the road from aerial images, with the goal of improving its performance and expanding its potential applications.

\subsection{Part 2 - Post processing}
As we mentioned in the last section, we evaluated the impact of three threshold values to the refined output:
\begin{enumerate}
\item Overlap percentage threshold: As we mentioned in the method section, one drawback of our proposed method is that as soon as the overlap percentage reaches the threshold, it will return the overlapped location, but the overlap location might just be part of the real line. Even though we later grouped all locations into different clusters and fixed this issue, there are still some improvements we can do. One potential improvement is that we can set the threshold for the overlap percentage higher to further improve the reliability of the results. In the future, we could also explore using more advanced algorithms, such as machine learning models or object detection techniques.

\item Row distance threshold: In our method, it relies on manual input to determine the row distance threshold. This means that our method may not be effective for complex roads where the row distance varies across the image. In order to improve the reliability of our method, we need to find a way to automatically determine the row distance threshold. This could involve using image processing techniques to analyze the input image and identify the row distance on different parts of the road, or using machine learning algorithms to automatically determine the optimal row distance threshold for each input image.

\item Column distance threshold: The column distance threshold in our method provides a reasonable level of accuracy since we are using the average distance to determine if two columns belong to one cluster. However, this approach is computationally intensive, and may not be suitable for applications that require real-time performance. In order to improve the efficiency of our method, we could explore using more advanced algorithms to automatically determine the optimal column distance threshold for each input image. This would allow us to achieve a similar level of accuracy without incurring the computational overhead of calculating the average distance between columns.

\end{enumerate}
\printbibliography 

\end{document}